\documentclass[conference]{IEEEtran}
\IEEEoverridecommandlockouts
\usepackage{cite}
\usepackage{amsmath,amssymb,amsfonts}
\usepackage[cmintegrals]{newtxmath}
\usepackage{graphicx}
\usepackage{float}
\usepackage{caption}
\usepackage{subcaption}
\usepackage{textcomp}
\usepackage{xcolor}
\usepackage{multirow}
\usepackage{booktabs}
\usepackage[ruled, vlined]{algorithm2e}

\allowdisplaybreaks
\def\BibTeX{{\rm B\kern-.05em{\sc i\kern-.025em b}\kern-.08em
    T\kern-.1667em\lower.7ex\hbox{E}\kern-.125emX}}
\begin{document}

\title{Feature-Attention Graph Convolutional Networks for Noise Resilient Learning}

\author{
\IEEEauthorblockN{Min Shi\textsuperscript{1}, Yufei Tang\textsuperscript{1}, Xingquan Zhu\textsuperscript{1} and Jianxun Liu\textsuperscript{2}}
\IEEEauthorblockA{
\textsuperscript{1} Department of Computer \& Electrical Engineering and Computer Science, Florida Atlantic University, USA \\
\textsuperscript{2} School of Computer Science and Engineering, Hunan University of Science and Technology, China
\\ Email: \{mshi2018, tangy, xzhu3\}@fau.edu, ljx529@gmail.com}
}

\maketitle

\begin{abstract}
Noise and inconsistency commonly exist in real-world information networks, due to inherent error-prone nature of human or user privacy concerns. To date, tremendous efforts have been made to advance feature learning from networks, including the most recent Graph Convolutional Networks (GCN) or attention GCN, by integrating node content and topology structures. However, all existing methods consider networks as error-free sources and treat feature content in each node as independent and equally important to model node relations. The erroneous node content, combined with sparse features, provide essential challenges for existing methods to be used on real-world noisy networks. In this paper, we propose FA-GCN, a feature-attention graph convolution learning framework, to handle networks with noisy and sparse node content. To tackle noise and sparse content in each node, FA-GCN first employs a long short-term memory (LSTM) network to learn dense representation for each feature. To model interactions between neighboring nodes, a feature-attention mechanism is introduced to allow neighboring nodes learn and vary feature importance, with respect to their connections. By using spectral-based graph convolution aggregation process, each node is allowed to concentrate more on the most determining neighborhood features aligned with the corresponding learning task. Experiments and validations, \textit{w.r.t.} different noise levels, demonstrate that FA-GCN achieves better performance than state-of-the-art methods on both noise-free and noisy networks.
\end{abstract}


\IEEEpeerreviewmaketitle

\section{Introduction}

\IEEEPARstart{M}{any} real-world applications involve knowledge mining and analysis from network or graph-based data such as citation networks, social networks, telecommunication networks, and biological networks, \textit{etc}, where data are often collected from noisy channels with erroneous/inconsistent labels or features~\cite{nt2019learning}. In order to carry out pattern mining from networks, such as community detection~\cite{wang2017community}, node classification~\cite{yao2018graph}, link prediction~\cite{zhang2018link}, \textit{etc.}, network representation learning (or embedding learning)~\cite{zhang2018network} is commonly used to construct features to represent nodes for learning.

To capture node relations, early network embedding learning mainly focuses on topology features~\cite{ribeiro2017struc2vec, wang2017community}, where nodes sharing similar topology structures are enforced to have close feature representation. For example, two scholarly publications citing same set of literature in a citation network would be represented by similar feature vectors \cite{le2014probabilistic}, and two users interacting with many common friends in a social network would share similar features in the learned representation space \cite{perozzi2014deepwalk}. However, structure-based methods can only model the explicit node relations already reflected by the network edges, which fail to capture the implicit relationships between nodes because of the sparse graph connections. For example, two users in a social network do not have an immediate link not because they are not friends in reality, but they might be unaware of each other's existence online. To mitigate this problem, recent studies propose to embed the content information associated with a network to enhance node structures modeling \cite{yang2015network, pan2016tri}. Indeed, networks with rich textual contents are ubiquitous in real world, such as the citation network and Wikipedia network where nodes are usually described by substantial texts. In general, content features are able to reveal relationships between nodes aligned with the network structures (\textit{e.g.}, two nodes with many shared content features are highly likely to form a neighborhood) \cite{chang2009relational, wang2014knowledge}, but in a more fine-grained and interpretable fashion, \textit{i.e.}, the affinity between two linked nodes can be measured by the number of shared content features other than just a single edge in the graph.

In additional to the above adjacency matrix or random walk based network embedding learning, recently, the spectral-based Graph Convolutional Networks (GCN) \cite{kipf2016semi, schlichtkrull2018modeling} have shown impressive performance to directly embed graphs with rich content features by a semi-supervised node classification training. GCN relies on the assumption that each node tends to have the same label with its neighbors that is guaranteed by the aggregated features from all neighborhood nodes, where different features are typically treated as independent and equally important. However, such a learning mechanism may be challenged by the following two realities. First, in graphs (e.g., citation networks and Wikipedia networks) where contents are word sequences, each word feature usually does not appear to represent a complete semantic alone but to correlate with others to form an unique meaning \cite{qu2005new}, \textit{i.e.}, the meaning of a sentence is usually demonstrated not by every single word, but by the context of all words having dependencies or correlations with each other. Second, while nodes have connections are assumed to have dependencies with their content features, not all features function equally to trigger interactions between nodes, \textit{i.e.}, although a research publication may have rich text information such as title and abstract, yet many words are actually not distinguishing features to reveal its citation relationships with others. 

Indeed, existing methods have made significant progress for network embedding learning, but they all consider networks as error-free sources and treat features in each node content as independent and equally important while modeling the node relations. The ignorance of the impact of erroneous content, combined with sparse node features, provide essential challenges for existing methods to handle real-world noisy networks. In fact, a recent work has empirically validated various embedding approaches on sparse and noisy knowledge graphs, and concluded that ``\textit{embeddings are sensitive to sparse and unreliable data}''~\cite{pujara2017sparsity}. 

In summary, existing methods are sensitive and ineffective to noise and sparse content mainly because of the following challenges.
\begin{itemize}
    \item \textbf{Noise Propagation:} When noise is imposed to the node (\textit{e.g.} incorrect words), it will force existing methods, such as GCN or attention networks, to learn deteriorated weight values, corresponding to noisy features. Such noise propagation directly deteriorates network embedding results as we will show in Section 5.
    \item \textbf{Feature Interaction:} While existing methods have taken node content into consideration, they treat all features equally for embedding learning. In reality, features have different interaction with respect to neighboring nodes, and should be differentiated for learning each node's representation. 
    \item \textbf{Sparsity and Dimensionality:} Most networks have high dimensionality and sparse node content, (\textit{e.g.} each node only has about 1\% features, compared to the whole feature space). Noise impact, in a high dimensionality and sparse content setting, is even more severe because the underlying models are highly vulnerable to errors. 
\end{itemize}

\begin{figure}
  \centering
    \includegraphics[width=0.4\textwidth]{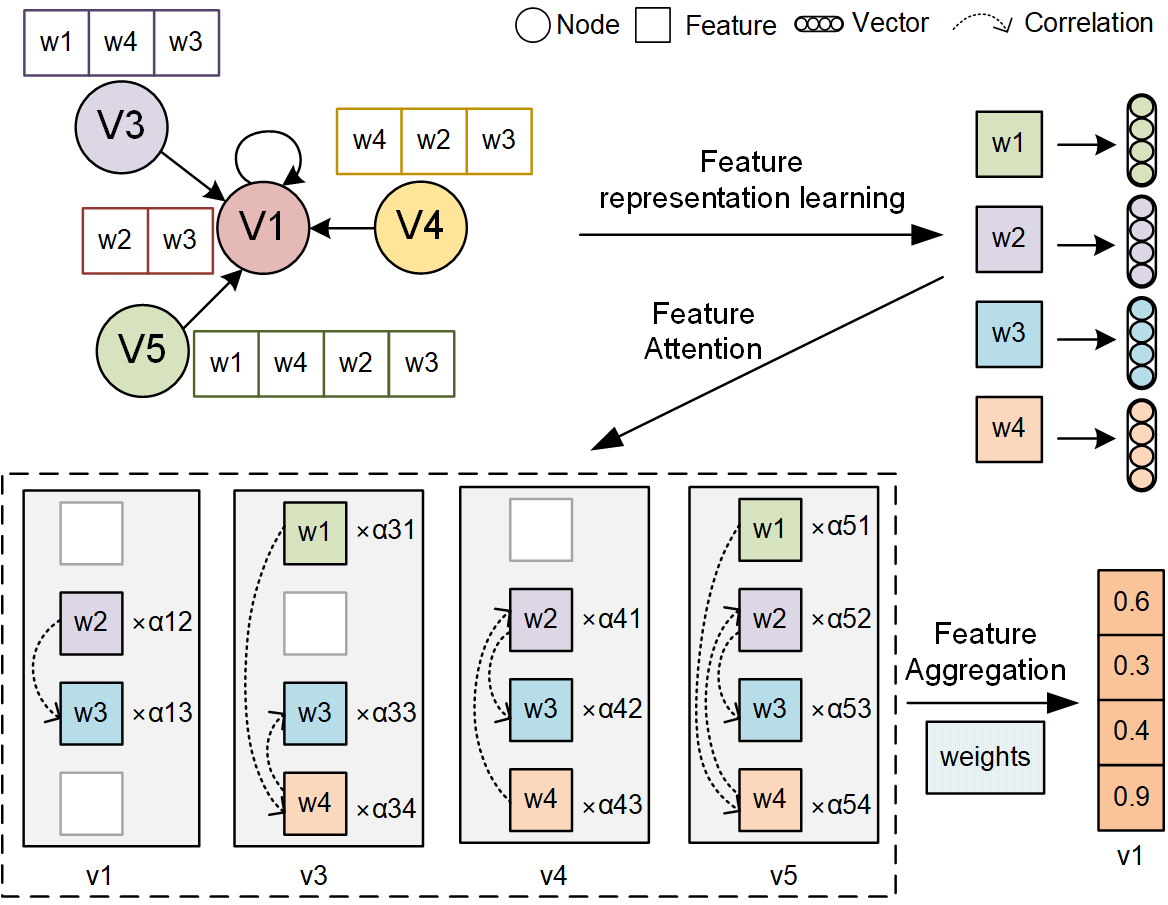}
  \caption{An illustrative example of the proposed approach: Feature representation is used to explore feature correlation and learn a dense vector for each feature. Feature attention is used to differentiate feature interaction between each node and its neighboring node features, allowing better feature aggregation for noise resilient embedding learning.}
  \label{fig:motivation}
\end{figure}

To address aforementioned problems, we propose a novel Feature-based Attention GCN (FA-GCN) model to perform noise resilient learning for networks with sparse and noisy node content. Figure \ref{fig:motivation} shows an illustrative example of our proposed approach, each content feature is first represented as a dense semantic vector, with feature correlation/dependency being well preserved based on a Bi-directional Long Short-Term Memory (LSTM) network. In other words, the representation learning for each feature is dependent on the semantic representations of other features in a node content. Meanwhile, to minimize the impact of noisy content features, we introduce an attention layer over the LSTM network to determine the importance of various neighborhood features, aligned with the corresponding node classification task. As a result, noisy features/words in a node will receive reduced attention values to minimize its impact, and resulting in noise-resilient learning.  


It is worth noting that our work is different from a recently proposed Graph Attention Networks (GAT) \cite{velivckovic2017graph}, wherein features are modeled as independent and the attention is calculated at the node level. In comparison, we argue that features in a node content could interact with each other to reveal richer and more accurate node semantics, \textit{i.e.}, the same word feature may have different meanings under different contexts and different word features may indicate the same meaning within a similar context. Meanwhile, the node-level attention in GAT assumes that all features in the node contents contributing equally to edge connections, whereas our feature-level attention enables differentiation of relevant features triggering node interactions in a network. Specifically, our main contributions are as follows:
\begin{itemize}
\item We proposed to model node relations at feature level, where each node interacts with different neighbors are attributed to the most influential node features.
\item We proposed to model feature correlations for enhanced node representation learning and classification, which is more reasonable for graphs where node contents or features are word sequences.
\item We proposed a noise-resilient learning framework for networks with sparse and noisy text features. It models feature correlations by a Bi-directional LSTM network and meanwhile conducts differentiable neighborhood features aggregation by a higher attention layer over the LSTM network.
\end{itemize}

The rest of the paper is organized as follows. Section 2 discusses related work, followed by problem definition and preliminaries related to LSTM network and spectral-based graph convolutional networks in Section 3. The proposed algorithm is described in Section 4, and experiments and comparisons are reported in Section 5. Finally, Section 6 concludes the paper.  

\section{Related Work}
Graph representation learning~\cite{zhang2018network, wu2019comprehensive} aims to represent each node of a target network as a low-dimensional vector, such that various downstream analytic tasks can be benefited. 
Early work in the area mainly focuses on shallow neural models to preserve only the node structures \cite{zhang2018network}. To capture high-order neighborhood relationships between nodes, DeepWalk \cite{perozzi2014deepwalk} performs a random walk process over the whole graph to generate  a collection of fixed-length node sequences similar to the natural language sentences. It then explores a widely used neural model Skip-Gram \cite{mikolov2013efficient} to learn node representations from these node sequences. However, Node2vec \cite{grover2016node2vec} demonstrates that DeepWalk has not fully preserved the connectivity patterns between nodes and thus proposes to combine the breadth-first sampling and depth-first sampling in the random walk process, where the community properties between nodes can be well preserved. LINE \cite{tang2015line} is proposed for large scale network representation learning by preserving the first- and second- order node relations, where the first-order is determined by the immediate links and the second-order relation between two nodes is created by their shared neighbors. However, in addition to the complex graph structures that have encoded node relations, graphs are usually associated with rich content information such as attributes and texts that also revealed the affinities between nodes \cite{le2014probabilistic, tu2017cane}. For examples, the Relational Topic Model (RTM) \cite{le2014probabilistic} is utilized to model both the documents and link relationships, which assumes that documents with links also have similar topic distributions and semantic representations. TADW \cite{yang2015network} leverages the rich texts to enhance the structure-based representation learning based on an equivalent matrix factorization method as the DeepWalk. TriDNR \cite{pan2016tri} can integrate the node structure, content and labels in an unified framework, which enforces the node representations to be learned from simultaneously the network structure and text content under the shared model parameters.

Because shallow models have limitations in learning complex relational patterns between nodes \cite{wu2019comprehensive}, there is an increasing number of efforts to explore graph neural networks, which take a graph as input and learn node representations by a supervised training process \cite{scarselli2009graph}. Recently, inspired by the huge success of convolutional neural networks on grid-like data such as images, a lot of tentative works emerged that seek to adopt a similar convolutional feature extraction process directly on the arbitrarily structured data such as graphs \cite{defferrard2016convolutional, bruna2013spectral}. To date, Graph Convolutional Networks (GCN) \cite{kipf2016semi} have appeared to achieve the state-of-the-art performance in many graph related analytic tasks, which can naturally learn node representations from graph structures and contents. For example, Schlichtkrull et al. \cite{schlichtkrull2018modeling} proposed the relational GCN and applied them to two standard knowledge base completion tasks of link prediction and entity classification. Yao et al. \cite{yao2018graph} proposed the Text GCN for text classification, where the text graph is built based on the word co-occurrence and document word relations. Yan et al. \cite{yan2018spatial} proposed the spatial-temporal GCN for skeleton-based action recognition. It is formulated on top of a sequence of skeleton graphs, where each node corresponds to a joint of the human body and  edges represent the connectivity between joints. In general, while a node could connect with many others in a graph, different neighbors may have different contributions when generating the representation of this node. The Graph Attention Networks \cite{velivckovic2017graph} were proposed to solve this problem by using a self-attention strategy to assign large weights on important neighbors for feature aggregations. However, since nodes interact with each other are usually resulted by fine-grained features \cite{tu2017cane}, node level attention maybe insufficient to characterize node relations. 

In comparison, all existing work considers networks as quality sources without considering noise or errors in the networks. As a recent study~\cite{pujara2017sparsity} has show that all embedding methods suffer significantly from ``sparse and unreliable data'',  we propose a feature-attention mechanism to differentiate and aggregate neighbor features for noise resilient learning. 

\section{Problem Definition \& Preliminaries}

\subsection{Problem Definition \& Motivation}
A network is represented as $G=(V,E,\mathbb{C})$, where ${V}=\left\{{v_i} \right\}_{i = 1,\cdots, n}$ is a set of unique nodes and ${E}=\{ e_{i,j} \}_{ {i,j}=1,\cdots,n;~i \neq j }$ is a set of edges. Let $A$ denote the $n \times n$ adjacency matrix representation of edges with $A_{i,j}=1$ if $e_{i,j}\in E$ and $A_{i,j}=0$ if $e_{i,j}\notin E$. Let $D$ and $I$ be the $n \times n$ degree matrix and identity matrix respectively, where $D_{ii}$ is calculated by $D_{ii}=\sum_{j}A_{ij}$. For each node $v_i$, we use $cnt_i$ to denote its content, which is a sequence of word features represented by $cnt_i=\left\{ w_j\right\}_{j=1,\cdots,|cnt_i|}$. For all nodes in $G$, their contents form the content corpus $\mathbb{C}=\left\{ {cnt_i} \right\}_{i = 1,...,n}$. We use $|\mathbb{C}|$ to denote vocabulary (or number of unique words) in the content corpus. It is common that each node has very sparse content, so $|cnt_i| \ll |\mathbb{C}|$.  

In this paper, we refer to noise as erroneous node content, where content of each node $cnt_i$ contains some errors (\textit{e.g.} erroneous feature values or words). Under the sparse and noisy node content setting, our \textbf{goal} is to learn good feature representation for each node in the network for classification.  

In order to tackle sparse and noisy node content, our motivation, as shown in Figure~\ref{fig:motivation}, is to employ an optimization approach to address the sparsity and errors: (1) learning a dense vector to represent each content word ($w_j$), and (2) using feature-attention to learn weight values for each word, based on node-node interaction, and then use feature-attention to aggregate neighboring nodes for noise resilient representation learning for each node. 

As shown in Figure ~\ref{fig:motivation}, node $v_1$ has three neighbors ($v_3$, $v_4$, and $v_5$), each containing some content words. Our first learning objective is to learn a dense feature vector for each word ($w_j$). Then feature-attention is used to learn weight values $\alpha_{ij}$ which quantify node $v_1$'s feature level interactions with respect to neighbor $v_i$ on word $w_j$. After that, feature aggregation is used to aggregate all $V_i$'s neighbors to learn a good feature representation for $v_i$. It is worth noting that the learning of feature/word representation and node representation are carried out simultaneously under a unified optimization goal, as we will detail in Eq.~(\ref{eq:objective}).

\subsection{Long Short-Term Memory}
To learn vector representation for words, Long Short-Term Memory (LSTM) networks \cite{hochreiter1997long} have achieved significant success~\cite{cheng2016long,yang2017attention}, thanks to its recurrent learning capacity on sequential data like text. In LSTM, features are not independently modeled but can interact with each other through the memory and state transmission mechanisms. When two reverse-order LSTM networks are combined, each feature is allowed to semantically correlate with any other feature within the same sequence. Figure \ref{fig:lstm} shows the interior structure of an LSTM unit/cell \cite{hochreiter1997long}, where $f_t$, $i_t$ and $o_t$ are the forget, input and output gates, respectively. $h_t$ denotes the cell output at time $t$, and $c_t$ is the global cell state that enables the sharing of different cell outputs throughout the LSTM network. Features are usually sequentially fed into the LSTM network, where the corresponding parameters for a feature at time $t$ are updated by:
\begin{equation}
f_t=\sigma(w_{xf}x_t+w_{hf}h_{t-1}+b_f)
\end{equation}
\begin{equation}
i_t=\sigma(w_{xi}x_t+w_{hi}h_{t-1}+b_i)
\end{equation}
\begin{equation}
g_t=\tanh(w_{xg}x_t+w_{hg}h_{t-1}+b_g)
\end{equation}
\begin{equation}
f_t=f_{t}c_{t-1}+i_tg_{t}
\end{equation}
\begin{equation}
o_t=\sigma(w_{xo}x_t+w_{ho}h_{t-1}+b_o)
\end{equation}
\begin{equation}
h_t=o_t\tanh(c_t)
\end{equation}
where $w_{x \ast }=\left \{ w_{xf},w_{xi},x_{xg},w_{xo} \right \}$ and $w_{h \ast }=\left \{ w_{hf},w_{hi},x_{hg},w_{ho} \right \}$ are weight parameters for the corresponding gates, and $b_f$, $b_i$, $b_g$ and $b_o$ are their biases, respectively.

\subsection{Graph Convolutional Networks}
GCN is an efficient variant of the convolutional neural networks operating directly on graphs \cite{kipf2016semi} by encoding both the graph structures and node features. Given a network $G=(V,E,X)$, which has $n$ vertices and each node has $d_o$ dimension feature values ($X\in \mathbb{R}^{n \times d_o}$ denotes the feature value matrix), GCN intends to learn a low-dimensional node representations though a convolutional learning process. In general, with one convolutional layer, GCN is able to preserve the first-order neighborhood relations between nodes, where each node is represented as an $l-$dimension vector, and node feature matrix $X^{(1)} \in \mathbb{R}^{n \times l}$ is computed by:
\begin{figure}
  \centering
    \includegraphics[width=0.30\textwidth]{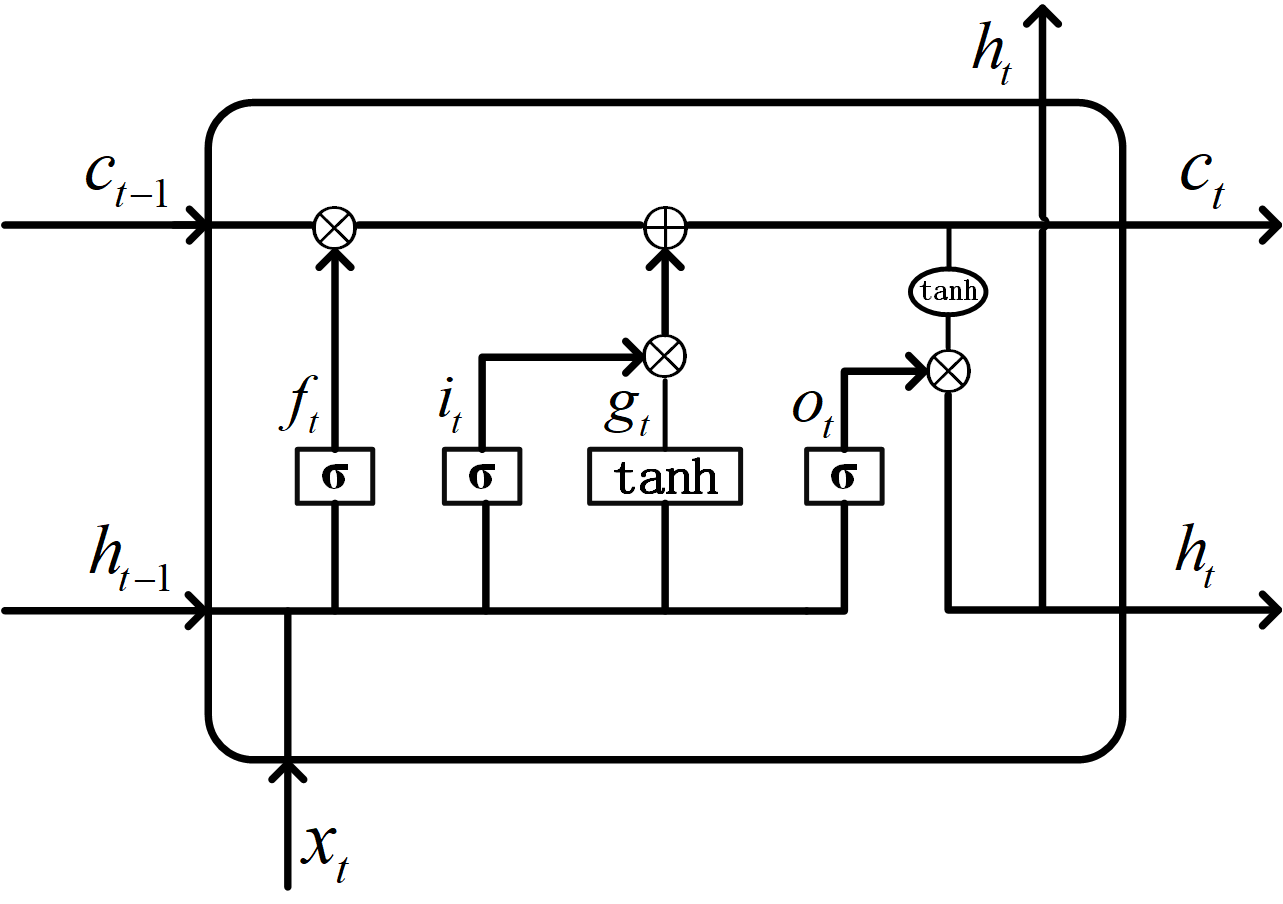}
  \caption{Structure of an LSTM unit/cell.}
  \label{fig:lstm}
 \vspace{-4mm}
\end{figure}

\begin{figure*}
\centering 
\begin{minipage}[b]{0.78\linewidth}
\includegraphics[width=1\textwidth]{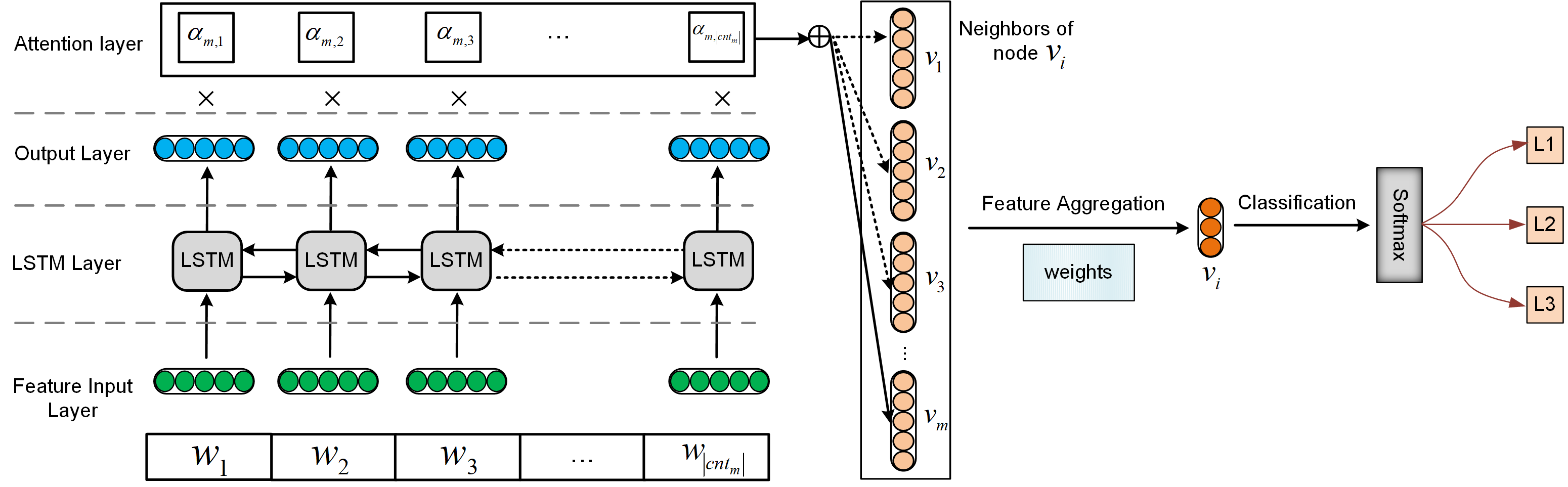}
\end{minipage}
\caption{The proposed FA-GCN model. Feature representations of nodes are dynamically learned by the LSTM network with attention mechanisms. Then, with the spectral-based convolutional filter, each node could gather the most important content features from itself and its neighbors to form the node representation. The feature and node representations are trained in an unified manner to optimize the collective classification objective by Eq. (28).}
\label{fig:framewrok}
\end{figure*}
\begin{equation}
X^{(1)} = \rho (\tilde{A} X^{(0)} W_{0})
\end{equation}
where $\tilde{A}=D^{-\frac{1}{2}}(I+A)D^{-\frac{1}{2}}$ is the normalized symmetric adjacency matrix, $X^{(0)} \in \mathbb{R}^{n \times d_o}$ is the initial input feature matrix of $X$ and $W_{0} \in \mathbb{R}^{d_o\times l}$ is a weight matrix for the first convolutional layer. $\rho$ is an activation function such as the \textit{ReLU} represented by $\rho (x) = \max (0,x)$. If it is necessary to encode higher-order (e.g., $k$-hop) neighborhood relationships, one can easily stack multiple GCN layers, where the output node features of the $j$th (0 $\leq$ $j$ $\leq$ $k$) layer is computed by:
\begin{equation}
X^{(j+1)} = \rho (\tilde{A} X^{(j)} W_{j})
\end{equation}
where $W_j \in \mathbb{R}^{d_o \times d_o}$ (or $W_j \in \mathbb{R}^{d_o \times l}$ if it is the last layer) is the weight matrix for the $j$th layer.

\section{The Proposed Approach}
For existing GCN-based methods,  they model node features as independent (\textit{e.g.} using one-hot representation of content features where each element in the feature matrix $X$ indicates whether a corresponding feature (\textit{e.g.}, word) appears or not), making these methods sensitive to noise and sparse input. 

In order to tackle high dimensional, sparse, noisy node content, we propose to represent each single node feature as a dense semantic vector and various features can influence each other by interactions of their semantic vectors during learning. In addition, each feature will be assigned with an importance weight, which allows each node aggregates important neighbor features for representation learning.

The proposed Feature Attention-based GCN (FA-GCN) model for above two purposes is shown in Figure \ref{fig:framewrok}. First, the feature representations are learned by a Bi-directional LSTM network, where feature correlations can be preserved. Then, with an attention layer on the top, each feature is weighted based on its importance to the target node classification task. In this paper, we investigate the effectiveness of two types of feature attention mechanisms that either consider a self-transformation process or introduce a context-aware bilinear term. Finally, each node dynamically aggregates weighted sum of neighborhood features to form its representation. The feature representations and node representations are integrally learned and could enhance each other to optimize a collective classification loss at the end of this framework.

\subsection{Feature Representation Learning}
Network node features contain rich semantics and they frequently correlate with each other to trigger complex node connections in a graph. For example, a word feature may have different semantics when correlating with different words under different sentence contexts and these fine-grained semantics could help differentiate the links of the corresponding node with its different neighbors. 

As shown in Figure \ref{fig:framewrok}, we use a Bi-directional LSTM network to learn representation of each feature from the content corpus $\mathbb{C}$. Let the content features of node $v_m$ be represented by  $cnt_m=\left\{ w_j\right\}_{j=1,\cdots,|cnt_m|}$, we initialize the representations (\textit{e.g.}, with dimension $d_{i}$) of these features in the input layer with following a uniform range distribution. Assume the input semantic vector for feature $w_j$ at time $t$ is represented by $vec_{w_j} \in \mathbb{R}^{d_i}$, it then undergoes a series of non-linear transformations in temporal order formulated by:
\begin{equation}
\overrightarrow{f}_{t,w_j}=\sigma(\textbf{W}_f\textbf{X}_t+b_f)
\end{equation}
\begin{equation}
\overrightarrow{i}_{t,w_j}=\sigma(\textbf{W}_i\textbf{X}_t+b_i)
\end{equation}
\begin{equation}
\overrightarrow{g}_{t,w_j}=\tanh(\textbf{W}_g\textbf{X}_t+b_g)
\end{equation}
\begin{equation}
\overrightarrow{c}_{t,w_j}=\overrightarrow{f}_{t,w_j}\overrightarrow{c}_{t-1,w_{j-1}}+\overrightarrow{i}_{t,w_j}\overrightarrow{g}_{t,w_j}
\end{equation}
\begin{equation}
\overrightarrow{o}_{t,w_j}=\sigma(\textbf{W}_o\textbf{X}_t+b_o)
\end{equation}
\begin{equation}
\overrightarrow{h}_{w_j}=\overrightarrow{o}_{t,w_j}\tanh(\overrightarrow{c}_{t,w_j})
\end{equation}
and
\begin{equation}
\textbf{W}_*=[w_{x*},w_{h*}], \textbf{X}_t=[vec_{w_j}, h_{w_{j-1}}]
\end{equation}
where $w_{j-1}$ represents the feature at time $(t-1)$. Then, we use the element-wise sum to combine outputs of feature $w_j$ from the two opposite LSTM layers to form its final semantic vector representation:
\begin{equation}
h_{w_{j}}=\overrightarrow{h}_{w_j} \oplus \overleftarrow{h}_{w_j}
\end{equation}
where $\overleftarrow{h}_{w_j} \in \mathbb{R}^{d_o}$ represents the vector output of $w_j$ from the  opposite temporal order, which is calculated in a similar way as Eq.(14).

\subsection{Feature-Attention Mechanisms}
Attention mechanisms have been widely used in many sequence-based tasks such as sentiment classification \cite{yang2017attention} and machine translation \cite{luong2015effective}, which are favorable designs allowing models to learn alignments between different modalities, \textit{i.e.}, focusing on the most relevant neighborhood features that are helpful to the node classification task. In this paper, we investigate two types of attention mechanisms both at the feature level. \\
\textbf{Attention 1} Let the semantic vectors of all features in node content $cnt_m$ be represented by $H_m=[h_{w_1},h_{w_2},\cdots,h_{w_{|cnt_m|}}]$, where $H_m \in \mathbb{R}^{|cnt_m| \times d_o}$ and $h_{w_{j=1,2,\cdots,|cnt_m|}}$ is calculated from Eq.(16). Inspired by a recent attention design that aims to capture the most important/relevant words in a given sentence for relation classification \cite{zhou2016attention}, we calculate a weight vector $\alpha_m=\left\{ \alpha_j \right\}_{j=1,\cdots,|cnt_m|}$ for all features in $H_m$ by:
\begin{equation}
    M = \tanh(H_m)
\end{equation}
\begin{equation}
    \alpha_m = softmax(MW_a^T)
\end{equation}
where ${\alpha}_{m,j}$ is the attention weight for feature $h_{w_j}$. $M \in \mathbb{R}^{|cnt_m| \times d_o}$ is the non-linear transformation of $H_m$, $W_a \in \mathbb{R}^{d_o}$ is a trained parameter vector shared across all nodes and $W_a^T$ is the transpose. 

\noindent\textbf{Attention 2} However, the above way of obtaining attention scores adopts a simple self-transformation without considering the neighborhood relationships. Therefore, we propose to perform a context-aware attention calculation, which determines the importance of neighborhood features by taking the corresponding context node into consideration. In Figure 3, assume all neighbors (each node also aggregates information from itself) of node $v_i$ are represented by a collection $\mathbb{N}_i=\left \{ v_m \right \}_{m=1,\cdots,|\mathbb{N}_i|}$ and they have a shared contextual semantic vector computed by the element-wise sum of all individual feature vectors for node content $cnt_i$:
\begin{equation}
    h_{context_i} = \sum\limits_{j=1}^{|cnt_i|}H_{i,j}
\end{equation}
where $H_i=[h_{w_1},h_{w_2},\cdots,h_{w_{|cnt_i|}}]$ are semantic vectors of all features of node $v_i$ calculated based on Eq. (16). Then, the weight score $\alpha_j$ for each feature $h_{w_j} \in H_m$ of the neighborhood node $v_m$ is computed by using a bilinear term:
\begin{equation}
    {\alpha}_{m,j} = softmax(h_{w_j}W_bh_{context_i}^T)
\end{equation}
where $W_b \in \mathbb{R}^{d_o \times d_o}$ is a trained parameter matrix and $h_{context_i}^T$ is a transpose.

By incorporating the attention weights, the final feature representation $X_m \in \mathbb{R}^{d_o}$ for neighborhood node $v_m$ is formed by a weighted sum of all corresponding individual features by:
\begin{equation}
    X_m = \sum\limits_{j = 1}^{|cnt_m|}{\alpha}_{m,j}H_{m,j}
\end{equation}

\subsection{Node Representation Learning}
The spectral-based convolutional filter \cite{kipf2016semi} is chosen as a key building component in our framework to gather features dynamically learned by the LSTM network for node representation learning, where each node only aggregates features from itself and all its first-order neighbors. In this paper, we adopts a two-layer convolutional node representation learning process, where the embedding $X^{(1)}_i \in \mathbb{R}^{d_h}$ for each node $v_i$ in the first layer is computed by:
\begin{equation}
    {X^{(1)}_i} = \sum\limits_{m \in \mathbb{N}_i} W_0X_m
\end{equation}
where $W_0 \in \mathbb{R}^{d_h \times d_o}$ is the weight matrix and $d_h$ is the dimension of node embeddings in the first layer. Assume embeddings of all nodes out in the first layer are represented by $X^{(1)}=\left\{ X^{(1)}_i \right\}_{i=1,\cdots,n}$, then the node embeddings in the second layer are computed by: 
\begin{equation}
    O=\tilde{A} ReLU(X^{(1)})W_1
\end{equation}
where $\tilde{A}=D^{-\frac{1}{2}}(I+A)D^{-\frac{1}{2}}$ is the normalized symmetric adjacency matrix with self-loops. $W_1 \in \mathbb{R}^{d_h \times l}$ is the parameter matrix that transforms each node embedding to a $l$-length vector. Finally, the output node embeddings $O \in \mathbb{R}^{n \times l}$ of the last layer are subsequently through a \textit{softmax} classifier to perform a multi-class classification task by:

\begin{equation}
    Z=softmax(O)=\frac{exp(O)}{\sum\nolimits_i exp(O_i)}
\end{equation}
Let $Y$ be the one-hot label indicator matrix of all nodes, the classification loss can be defined as the cross-entropy error by:
\begin{equation}
  \mathcal{L}=-\sum\limits_{d \in \mathcal{Y}_L}\sum\limits_{f = 1}^l {Y_{df} \ln Z_{df} }
\end{equation}
where $\mathcal{Y}_L$ is the set of node indices that have labels. 

\begin{algorithm}[t]
\small
\caption{FA-GCN: Feature-attention GCN for noise resilient learning}\label{EMDalg}
\SetAlgoLined
\SetKwInOut{Input}{Input}\SetKwInOut{Output}{Output}
\Input{An information network: $G=(V,E,\mathbb{C})$}
\Output{The node embeddings: $O \in \mathbb{R}^{n \times l}$}
\textbf{Initialization}: $i=0$, training epochs $I$ and labeled nodes $\mathcal{Y}_L$
\BlankLine
\While{$i \le I$}{
    \For{a vertex $v_{m} \in V$}{
        $H_m \leftarrow$ Learn the dense semantic vector for each feature in $cnt_m$ by Eq. (16); \\
        $X_m \leftarrow$ Learn original node feature representation by Eq. (21).
        }
    $O \in \mathbb{R}^{n \times l} \leftarrow$ Learn node embeddings by Eq. (23); \\
    $\mathcal{L}\leftarrow$ Calculate classification loss by Eq. (28); \\
    $\big[W_*, W_b, W_0, W_1\big]\leftarrow$ Update feature representation learning weights $W_*$ in Eq. (15), attention weight $W_b$ in Eq. (20), and node embedding learning weights $W_0$ \& $W_1$ in Eqs. (22) and (23); \\
    $i$ = $i$ +1.
}
\end{algorithm}

The weight parameters for feature representation learning (\textit{e.g.}, $w_{x \ast }=\left \{ w_{xf},w_{xi},x_{xg},w_{xo} \right \}$ and $w_{h \ast }=\left \{ w_{hf},w_{hi},x_{hg},w_{ho} \right \}$) and node representation learning (e.g., $W_0$ and $W_1$) are trained collectively using the gradient descent algorithm as in \cite{kipf2016semi} and \cite{yao2018graph}. Since the feature representation learning of one node can be influenced by that of other neighborhood nodes, both the LSTM network parameters and GCN network parameters could vary dramatically without regularization, which might leads to the over-fitting and instability problems. To mitigate these issues, we add a L2-norm regulation term on the loss function by:
\begin{equation}
    R_f^{\left \{ 2 \right \}} =  \sum\limits_{w_x \in w_{x*}} \|{w_x}\|_F^2 + \sum\limits_{w_h \in w_{h*}} \|{w_h}\|_F^2
\end{equation}
\begin{equation}
    R_n^{\left \{ 2 \right \}} = \|{W_0}\|_F^2+ \|{W_1}\|_F^2
\end{equation}
\begin{equation}
  \mathcal{L}=-\sum\limits_{d \in \mathcal{Y}_L}\sum\limits_{f = 1}^l {Y_{df} \ln Z_{df} } + \ \lambda_1R_f^{\left \{ 2 \right \}} + \lambda_2R_n^{\left \{ 2 \right \}}\label{eq:objective}
\end{equation}
where $\lambda_1$ and $\lambda_2$ are penalty terms to control the weight magnitude of the regularization terms on feature and node representation learning weight parameters, respectively. The collective learning process of FA-GCN is summarized in Algorithm 1.

\section{Experiments}
\subsection{Datasets}
We choose three widely used benchmark datasets described as follows: \\
\textbf{Citeseer} dataset contains 3,312 literature from 6 categories and 4,732 links between them. Each publication is described by a text with average number of words of 32, where the word features in each node content are not ordered in a meaningful sequence (e.g., alphabetical order). There are 3,703 unique words in the vocabulary.\\
\textbf{Cora} dataset contains 2,708 research papers from 7 machine learning directions such as \textit{Reinforcement Learning} and \textit{Genetic Algorithms}. Each paper corresponds to a category label. There are 5,214 citation relations between these papers. Each paper is described by an abstract in the form of word sequence. There are 14,694 unique words in the vocabulary and the average number of words for each node is 90. \\
\textbf{DBLP} dataset contains 10,310 publications from 4 research areas in computer science, including \textit{database}, \textit{data mining}, \textit{artificial intelligence} and \textit{computer vision}. There are 52,890 edges in total and each publication is associated with a title in the form of word sequence. There are 15,135 unique words in the vocabulary and the average number of words for each publication is 8.

\begin{table}
\small
\centering
\caption{Benchmark network characteristics.}
\label{tab:data}
    \begin{tabular}{c|c|c|c}
    \hline
    Items & Citeseer & Cora & DBLP  \\
    \hline
    \# Nodes & $3,312$ & $2,211$  & $17,725$  \\
    \hline
    \# Edges & $4,732$ & $5,001$ & $52,890$ \\
    \hline
    \# unique words & $3,703$ & $9,679$ & $6,974$ \\
    \hline
    \# average words per node & $32$ & $87$ & $7$ \\
    \hline
    \# Categories & $6$ & $7$ & $4$ \\
    \hline
    \end{tabular}
\vspace{-3mm}
\end{table}

The detailed statistic information is summarized in Table 1. It is necessary to mention that since both Cora and DBLP have sequential word features, it is reasonable to consider the feature dependencies or correlations for more accurate relationship modeling between nodes. In addition, to further evaluate the capacity of the proposed approach to model feature correlation and feature attention in a more general setting, we also use the Citeseer dataset in which features are disordered, \textit{i.e.}, features of all nodes are sorted by the alphabetical order. As we adopt a Bi-directional LSTM network to learn the feature representations, each feature is able to reach and interact with others within the same node content.

\begin{table*}[ht]
\renewcommand{\arraystretch}{1.10}
\small
\centering
\caption{Node classification results on Citeseer ($p\%$ denotes percentage of labeled nodes).}
\label{tab:results}
\begin{tabular}{c|c|c|c|c|c|c|c|c|c|c}
\toprule

\multicolumn{2}{c|}{$p\%$} & 10\% & 15\% & 20\% & 25\% & 30\% & 35\% & 40\% & 45\% & 50\%\\
\hline
\multirow{8}{4em}{Methods} & DeepWalk & {$28.34${\tiny $\pm1.01$}} & {$30.17${\tiny $\pm0.86$}} &   {$31.04${\tiny $\pm0.95$}} & {$32.17${\tiny $\pm0.80$}} & {$32.73${\tiny$\pm1.03$}} &  {$33.88${\tiny$\pm0.79$}} & {$34.13${\tiny$\pm0.92$}} &  {$34.58${\tiny$\pm0.96$}} &  {$35.08${\tiny$\pm0.64$}} \\

& Node2vec & {$47.98${\tiny $\pm1.75$}} & {$51.16${\tiny $\pm1.21$}} &   {$52.66${\tiny $\pm1.12$}} & {$53.75${\tiny $\pm0.85$}} & {$54.58${\tiny$\pm0.84$}} &  {$55.68${\tiny$\pm0.83$}} & {$56.24${\tiny$\pm0.75$}} &  {$56.81${\tiny$\pm1.21$}} &  {$56.58${\tiny$\pm0.96$}} \\

& TriDNR & {$49.45${\tiny $\pm0.84$}} & {$51.95${\tiny $\pm0.29$}} &   {$53.40${\tiny $\pm0.92$}} & {$55.31${\tiny $\pm0.49$}} & {$55.20${\tiny$\pm0.57$}} &  {$55.18${\tiny$\pm0.84$}} & {$56.79${\tiny$\pm0.57$}} &  {$57.14${\tiny$\pm1.01$}} &  {$57.13${\tiny$\pm0.56$}} \\

& GCN & {$\mathbf{71.06}${\tiny $\pm\mathbf{0.47}$}} & {$71.43${\tiny $\pm0.55$}} &   {$72.25${\tiny $\pm0.45$}} & {$73.88${\tiny $\pm0.70$}} & {$74.15${\tiny$\pm0.64$}} &  {$74.87${\tiny$\pm0.53$}} & {$76.57${\tiny$\pm0.55$}} &  {$76.82${\tiny$\pm0.46$}} &  {$77.45${\tiny$\pm0.65$}} \\

& GAT & {$\textit{70.42}${\tiny $\pm\textit{0.52}$}} & {$\mathbf{72.71}${\tiny $\pm\mathbf{0.47}$}} &   {$\textit{73.05}${\tiny $\pm\textit{1.02}$}} & {$74.36${\tiny $\pm0.70$}} & {$74.65${\tiny$\pm0.77$}} &  {$76.10${\tiny$\pm0.46$}} & {$77.15${\tiny$\pm0.38$}} &  {$77.60${\tiny$\pm0.51$}} &  {$78.47${\tiny$\pm0.64$}} \\

& FA-GCN$_{cor}$ & \underline{{$67.96${\tiny $\pm0.57$}}} & \underline{{$71.91${\tiny $\pm0.56$}}} &   {$72.36${\tiny $\pm0.49$}} & \underline{{$74.70${\tiny $\pm0.30$}}} & \underline{{$75.38${\tiny$\pm0.55$}}} &  \underline{{$76.64${\tiny$\pm0.63$}}} & \underline{{$79.32${\tiny$\pm0.69$}}} &  \underline{{$80.83${\tiny$\pm0.47$}}} &  \underline{{$81.06${\tiny$\pm0.57$}}} \\

& FA-GCN$_{self}$ & {$68.23${\tiny $\pm0.76$}} & {$\textit{72.07}${\tiny $\pm\textit{0.53}$}} &   \underline{{$72.88${\tiny $\pm0.52$}}} & {$\textit{75.04}${\tiny $\pm\textit{0.43}$}} & {$\textit{75.45}${\tiny$\pm\textit{0.82}$}} &  {$\textit{77.06}${\tiny$\pm\textit{0.62}$}} & {$\textit{79.60}${\tiny$\pm\textit{0.52}$}} &  {$\textit{81.29}${\tiny$\pm\textit{0.55}$}} &  {$\textit{81.66}${\tiny$\pm\textit{0.64}$}} \\

& FA-GCN & {$67.89${\tiny $\pm0.48$}} & {$71.47${\tiny $\pm0.59$}} &   {$\mathbf{73.09}${\tiny $\pm\mathbf{0.52}$}} & {$\mathbf{75.38}${\tiny $\pm\mathbf{0.60}$}} & {$\mathbf{76.89}${\tiny$\pm\mathbf{0.60}$}} &  {$\mathbf{77.50}${\tiny$\pm\mathbf{0.41}$}} & {$\mathbf{80.39}${\tiny$\pm\mathbf{0.60}$}} &  {$\mathbf{81.43}${\tiny$\pm\mathbf{0.58}$}} &  {$\mathbf{82.38}${\tiny$\pm\mathbf{0.46}$}} \\
\bottomrule
\end{tabular}

\end{table*}

\begin{table*}[ht]
\small
\centering
\caption{Node classification results on Cora ($p\%$ denotes percentage of labeled nodes).}
\label{tab:results}
\begin{tabular}{c|c|c|c|c|c|c|c|c|c|c}
\toprule

\multicolumn{2}{c|}{$p\%$} & 10\% & 15\% & 20\% & 25\% & 30\% & 35\% & 40\% & 45\% & 50\%\\
\hline
\multirow{8}{4em}{Methods} & DeepWalk & {$43.42${\tiny $\pm1.53$}} & {$45.74${\tiny $\pm0.96$}} &   {$48.95${\tiny $\pm0.96$}} & {$50.22${\tiny $\pm1.02$}} & {$51.18${\tiny$\pm0.97$}} &  {$52.23${\tiny$\pm1.40$}} & {$53.23${\tiny$\pm1.10$}} &  {$53.34${\tiny$\pm1.01$}} &  {$54.40${\tiny$\pm1.35$}} \\

& Node2vec & {$68.66${\tiny $\pm1.83$}} & {$71.19${\tiny $\pm1.78$}} &   {$73.51${\tiny $\pm1.50$}} & {$74.79${\tiny $\pm1.35$}} & {$76.36${\tiny$\pm1.01$}} &  {$77.10${\tiny$\pm1.05$}} & {$78.03${\tiny$\pm0.95$}} &  {$78.43${\tiny$\pm1.33$}} &  {$78.78${\tiny$\pm1.12$}} \\

& TriDNR & {$74.32${\tiny $\pm1.53$}} & {$75.31${\tiny $\pm0.96$}} &   {$75.67${\tiny $\pm0.79$}} & {$75.19${\tiny $\pm0.64$}} & {$76.01${\tiny$\pm1.21$}} &  {$77.83${\tiny$\pm0.79$}} & {$78.45${\tiny$\pm0.96$}} &  {$78.36${\tiny$\pm0.96$}} &  {$79.20${\tiny$\pm0.79$}} \\

& GCN & {$\textit{82.13}${\tiny $\pm\textit{0.48}$}} & {$\textit{82.68}${\tiny $\pm\textit{0.52}$}} &   {$84.14${\tiny $\pm0.80$}} & {$84.68${\tiny $\pm0.41$}} & {$85.42${\tiny$\pm0.65$}} &  {$85.40${\tiny$\pm0.40$}} & \underline{{$87.34${\tiny$\pm0.35$}}} &  {$85.25${\tiny$\pm0.65$}} &  {$87.11${\tiny$\pm0.65$}} \\

& GAT & {$\mathbf{82.60}${\tiny $\pm\mathbf{0.47}$}} & {$\mathbf{83.34}${\tiny $\pm\mathbf{0.64}$}} &   {$\textit{84.33}${\tiny $\pm\textit{0.48}$}} & {$\textit{85.15}${\tiny $\pm\textit{0.39}$}} & \underline{{$86.23${\tiny$\pm0.95$}}} &  {$\textit{86.73}${\tiny$\pm\textit{0.72}$}} & {$\textit{87.62}${\tiny$\pm\textit{0.41}$}} &  \underline{{$87.44${\tiny$\pm0.81$}}} &  \underline{{$87.34${\tiny$\pm0.50$}}} \\

& FA-GCN$_{cor}$ & \underline{{$79.82${\tiny $\pm0.57$}}} & {$80.95${\tiny $\pm0.42$}} &   {$83.33${\tiny $\pm0.53$}} & {$84.29${\tiny $\pm0.44$}} & {$84.60${\tiny$\pm0.33$}} &  {$85.68${\tiny$\pm0.54$}} & {$86.77${\tiny$\pm0.57$}} &  {$87.07${\tiny$\pm0.39$}} &  \underline{{$87.34${\tiny$\pm0.71$}}} \\

& FA-GCN$_{self}$ & {$79.20${\tiny $\pm0.48$}} & {$82.21${\tiny $\pm0.54$}} &   \underline{{$84.24${\tiny $\pm0.37$}}} & \underline{{$84.84${\tiny $\pm0.61$}}} & {$\textit{86.33}${\tiny$\pm\textit{0.53}$}} &  \underline{{$86.45${\tiny$\pm0.60$}}} & {$87.32${\tiny$\pm0.45$}} &  {$\mathbf{87.82}${\tiny$\pm0.55$}} &  {$\textit{87.71}${\tiny$\pm\textit{0.77}$}} \\

& FA-GCN & {$78.20${\tiny $\pm0.42$}} & \underline{{$82.49${\tiny $\pm0.46$}}} &   {$\mathbf{84.50}${\tiny $\pm\mathbf{0.72}$}} & {$\mathbf{85.21}${\tiny $\pm\mathbf{0.38}$}} & {$\mathbf{86.96}${\tiny$\pm\mathbf{0.62}$}} &  {$\mathbf{87.56}${\tiny$\pm\mathbf{0.38}$}} & {$\mathbf{87.75}${\tiny$\pm\mathbf{0.48}$}} &  {$\textit{87.73}${\tiny$\pm\textit{0.51}$}} &  {$\mathbf{88.16}${\tiny$\pm0.61$}} \\
\bottomrule
\end{tabular}

\end{table*}

\subsection{Baselines}
We choose the following state-of-the-art comparative methods classified into three categories as follows.

\vspace{2pt}
\textbf{Structure only:}

\begin{itemize}
\item \textbf{DeepWalk} \cite{perozzi2014deepwalk} preserves only the neighborhood relations between nodes by the truncated random walk, and uses SkipGram model to learn the node embeddings.
\item \textbf{Node2vec} \cite{grover2016node2vec} adopts a more flexible neighborhood sampling process than DeepWalk, \textit{i.e.}, biased random walk, to better capture the local structure (the second-order node proximity) and the global structure (the high-order node proximity).
\end{itemize}

\vspace{2pt}
\textbf{Both structure and content without attention:}
\begin{itemize}
\item \textbf{TriDNR} \cite{pan2016tri} is a method that exploits network structure, node content and label information for node representation learning. It is based on the assumption that network structures and contents can enhance each other to collectively determine the affinities between nodes.
\item \textbf{GCN} \cite{kipf2016semi} is a state-of-the-art method that can efficiently model node relations from network structures and contents, where each node generate representation by adopting a spectral-based convolutional filter to recursively aggregate information from all its neighbors.
\item \textbf{FA-GCN$_{cor}$} is a variant of our proposed method that models the feature correlations by a Bi-directional LSTM network and then learns node representations based on the graph convolutional filter as GCN. The attention mechanism is not incorporated in this model.
\end{itemize}
\vspace{2pt}
\textbf{Both structure and content with attention:}
\begin{itemize}
\item \textbf{GAT} \cite{velivckovic2017graph} is a method built on the GCN model. It introduces an attention mechanism at the node level, which allows each node specifies different weights to different nodes in a neighborhood.
\item \textbf{FA-GCN$_{self}$} is a variant of our proposed method that learns feature representation based on the Bi-directional LSTM network with introducing a self-transformation attention mechanism (attention1 described in Section 4.2). It then learns node representations based on the graph convolutional filter.
\item \textbf{FA-GCN} is our proposed method that learns feature representation based on the Bi-directional LSTM network and learns node representations based on the graph convolutional filter. The context-aware attention mechanism (attention 2 proposed in Section 4.2) is adopted in this method.
\end{itemize}

\begin{table*}[ht]
\centering
\small
\caption{Classification results on DBLP ($p\%$ denotes percentage of labeled nodes).}
\label{tab:results}
\begin{tabular}{c|c|c|c|c|c|c|c|c|c|c}
\toprule

\multicolumn{2}{c|}{$p\%$} & 10\% & 15\% & 20\% & 25\% & 30\% & 35\% & 40\% & 45\% & 50\%\\
\hline
\multirow{8}{4em}{Methods} & DeepWalk & {$43.12${\tiny $\pm0.43$}} & {$43.91${\tiny $\pm0.41$}} &   {$44.42${\tiny $\pm0.45$}} & {$44.73${\tiny $\pm0.36$}} & {$44.98${\tiny $\pm0.46$}} & {$45.17${\tiny$\pm0.31$}} &  {$45.45${\tiny$\pm0.36$}} & {$45.41${\tiny$\pm0.32$}} &  {$45.57${\tiny$\pm0.37$}} \\

& Node2vec & {$76.08${\tiny $\pm0.39$}} & {$76.99${\tiny $\pm0.29$}} &   {$77.40${\tiny $\pm0.28$}} & {$77.56${\tiny $\pm0.27$}} & {$77.75${\tiny$\pm0.36$}} &  {$77.92${\tiny$\pm0.25$}} & {$77.96${\tiny$\pm0.36$}} &  {$78.02${\tiny$\pm0.26$}} &  {$78.16${\tiny$\pm0.30$}} \\

& TriDNR & {$49.45${\tiny $\pm0.84$}} & {$51.95${\tiny $\pm0.29$}} &   {$53.40${\tiny $\pm0.92$}} & {$55.31${\tiny $\pm0.49$}} & {$55.20${\tiny$\pm0.57$}} &  {$55.18${\tiny$\pm0.84$}} & {$56.79${\tiny$\pm0.57$}} &  {$57.14${\tiny$\pm1.01$}} &  {$57.13${\tiny$\pm0.56$}} \\

& GCN & {$\mathbf{82.89}${\tiny $\pm\mathbf{0.54}$}} & {$\mathbf{83.43}${\tiny $\pm\mathbf{0.59}$}} &   \underline{{$83.66${\tiny $\pm0.48$}}} & {$83.76${\tiny $\pm0.68$}} & {$84.31${\tiny$\pm0.85$}} &  {$84.29${\tiny$\pm0.40$}} & {$85.18${\tiny$\pm0.56$}} &  {$85.40${\tiny$\pm0.43$}} &  {$\mathbf{88.39}${\tiny$\pm\mathbf{0.88}$}} \\

& GAT & {$\textit{82.07}${\tiny $\pm\textit{0.39}$}} & \underline{{$82.71${\tiny $\pm0.50$}}} &   {$82.91${\tiny $\pm0.37$}} & {$83.36${\tiny $\pm0.47$}} & \underline{{$84.42${\tiny$\pm0.64$}}} &  {$\textit{85.72}${\tiny$\pm0.59$}} & {$\textit{86.15}${\tiny$\pm0.45$}} &  \underline{{$86.70${\tiny$\pm0.68$}}} &  {$87.42${\tiny$\pm0.60$}} \\

& FA-GCN$_{cor}$ & {$79.85${\tiny $\pm0.67$}} & {$82.16${\tiny $\pm0.83$}} &   {$83.62${\tiny $\pm0.51$}} & \underline{{$83.84${\tiny $\pm0.48$}}} & {$84.51${\tiny$\pm0.38$}} &  {$85.01${\tiny$\pm0.58$}} & \underline{{$85.87${\tiny$\pm0.43$}}} &  {$86.50${\tiny$\pm0.48$}} &  \underline{{$87.76${\tiny$\pm0.66$}}} \\

& FA-GCN$_{self}$ & \underline{{$80.95${\tiny $\pm0.50$}}} & {$82.45${\tiny $\pm0.57$}} &   {$\textit{83.87}${\tiny $\pm\textit{0.40}$}} & {$\textit{84.21}${\tiny $\pm\textit{0.70}$}} & {$\textit{85.15}${\tiny$\pm\textit{0.46}$}} &  \underline{{$85.16${\tiny$\pm0.68$}}} & {$85.85${\tiny$\pm0.50$}} &  {$\textit{86.93}${\tiny$\pm\textit{0.57}$}} &  {$87.53${\tiny$\pm0.61$}} \\

& FA-GCN & {$80.50${\tiny $\pm0.44$}} & {$\textit{82.87}${\tiny $\pm0.61$}} &   {$\mathbf{83.95}${\tiny $\pm\mathbf{0.48}$}} & {$\mathbf{84.51}${\tiny $\pm\mathbf{0.43}$}} & {$\mathbf{85.75}${\tiny$\pm\mathbf{0.51}$}} &  {$\mathbf{86.49}${\tiny$\pm\mathbf{0.57}$}} & {$\mathbf{86.82}${\tiny$\pm\mathbf{0.66}$}} &  {$\mathbf{87.63}${\tiny$\pm\mathbf{0.53}$}} &  {$\textit{88.24}${\tiny$\pm\textit{0.59}$}} \\
\bottomrule
\end{tabular}

\end{table*}

\begin{figure*}[t]
\centering 
\begin{minipage}[b]{0.25\linewidth}
\includegraphics[width=1\textwidth]{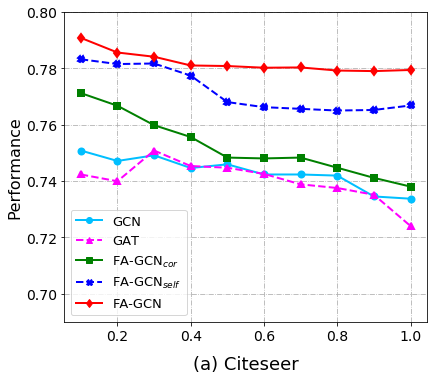}
\end{minipage}%
\hspace{1mm}
\begin{minipage}[b]{0.25\linewidth}
\centering
\includegraphics[width=1\textwidth]{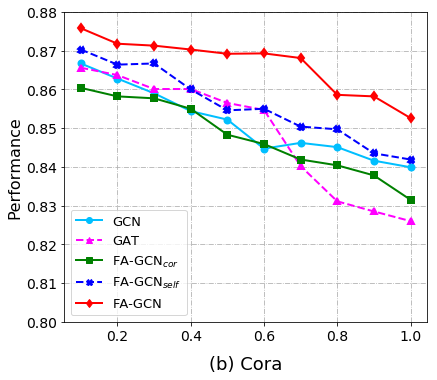}
\end{minipage}
\hspace{1mm}
\begin{minipage}[b]{0.25\linewidth}
\centering
\includegraphics[width=1\textwidth]{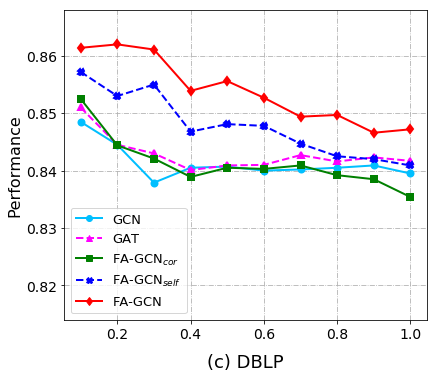}
\end{minipage}
\caption{Algorithm performance comparisons \textit{w.r.t.} different levels of injected noisy content features. The $x-$axis denotes noise levels, where 0.1 means adding 10\% of random noise to each node (\textit{e.g.} a node with 10 words would be injected one random word as noisy node content).}
\end{figure*}

\subsection{Experimental Settings}
\noindent \textbf{Node Classification.} We first perform the supervised node classification based on the learned node representations, which is a widely used way to demonstrate the graph learning performance \cite{perozzi2014deepwalk, pan2016tri, kipf2016semi}. $p \%$ labeled nodes are randomly selected for training the model (e.g., classifier), which is then used to predict labels for the rest of nodes. Similar to literature \cite{kipf2016semi, velivckovic2017graph}, we adopt \textit{Accuracy} to measure the classification performance, where experiments are repeated 5 times w.r.t different portions of training data and the average performance and standard deviation are finally reported.

\noindent \textbf{Noise Intervention.} We further test the performance of the proposed models to handle sparse and noisy content networks against various baselines. Two different types of noise intervention methods on the original node contents are adopted: 1) Inject different ratios (e.g., ranging from 0.1 to 1.0) of random noisy features into each node content; 2) Replace different ratios (e.g., between 0.05 and 0.5) of original features of each node content with randomly sampled noisy features. It is necessary to mention that the first intervention method will makes each node content contain more irrelevant features, while the second method will make each node content more erroneous and meanwhile become more sparser (e.g., the original correct content features are removed). Since the contents for DBLP network are already very sparse (e.g., 7 words per node), we only present the impacts of the second noise intervention w.r.t Citeseer and Cora datasets.

\noindent \textbf{Parameter Setting.} Extensive experiments are designed to test the sensitivities of various parameters. We test the input and out feature representation dimensions, $d_i$ and $d_o$, in LSTM network between 20 and 200, the training ratio, $p\%$, of the labeled nodes between 0.1 and 0.5. For comparison, the default settings for $d_i$, $d_o$ and $p\%$ are 80, 80 and 0.4, respectively. $d_h$ for Citeseer, Cora and DBLP are set as 6, 7 and 4, respectively. Both the LSTM and GCN networks use the dropout technique to reduce the effect of over-fitting, where dropout probabilities for LSTM and GCN are 0.2 and 0.3, respectively. The L2 norm regularization weight decay parameters $\lambda_1$ and $\lambda_2$ are respectively set as 5e-4 and 5e-4 for DBLP, and 5e-3 and 5e-4 for other datasets. We use the Adam optimizer to train the model, where the learning rate and training epoch are set as 2e-3 and 200, respectively.

\subsection{Experimental Results}

\noindent \textbf{Node Classification Performance.} Table 2, Table 3 and Table 4 presents the classification accuracy of all baselines on the three datasets, where the top three best results are bold-faced, italic-formatted and underscored, respectively. From the results, we have the following four main observations:
\begin{itemize}[leftmargin=10pt, topsep=4pt]
    \item From Table 2 and Table 3, we can conclude that methods incorporating both network structures and contents perform generally better than methods preserving only structures. For example, after modeling the text content of Cora network, the average performance of TriDNR improved 52.5\% over DeepWalk and 2.0\% over Node2vec, respectively. The reason is probably because of the fact that both Citeseer and Cora are sparse networks, where structures fail to capture the holistic relations between nodes. In such case, the rich network contents may can be leveraged to enhance the node relationship modeling. Above phenomenon can be strengthen by the results from Table 4 showing Node2vec performs significantly better than TriDNR, where the DBLP network has denser connectivity and sparser content compared with Citeseer and Cora networks. One the other hand, the shallow models such as DeepWalk, Node2vec and TriDNR suffers from the limitations of modeling complex node relations \cite{zhang2018network}, \textit{i.e}, they all use random walk over the network to capture the node relationships, but the random walk techniques cannot differentiate the affinities of node neighborhood relations of varying hops. In comparison, other GCN-based methods (e.g., GCN, GAT and FA-GCN$_{cor}$) can enforce a rigid neighborhood relations modeling by the efficient spectral-based convolutional feature aggregation process, where the learned node representations can naturally and precisely preserve the network structures and contents.

\begin{figure}[t]
\begin{minipage}[b]{0.5\linewidth}
\centering 
\includegraphics[width=1\textwidth]{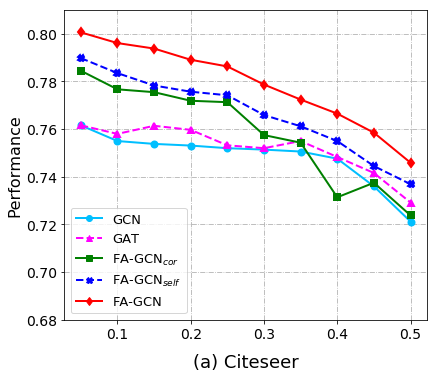}
\end{minipage}%
\begin{minipage}[b]{0.5\linewidth}
\centering
\includegraphics[width=1\textwidth]{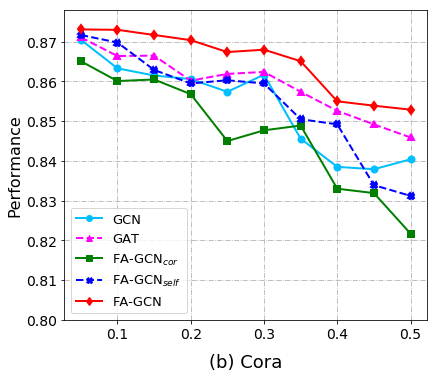}
\end{minipage}
\caption{Algorithm performance comparisons \textit{w.r.t.} different levels of replaced noisy content features. The $x-$axis denotes noise levels, where 0.1 means replacing 10\% of words in each node as noise (\textit{e.g.} a node with 10 words would have one word being replaced with a random word).}
\end{figure}
    
    \item Based on the classification results over all three datasets, in most cases FA-GCN$_{cor}$ outperforms GCN, \textit{i.e.}, an improvement of 1.7\% w.r.t the Citeseer dataset. The superiority was actually brought by a more reasonable way of modeling the network content features in the proposed model. Existing GCN-based methods typically take the static content features as input, where features are treated as independent and nodes link each others are assumed to have dependencies with their shared individual features. However, in many situations (e.g., especially for text-described networks) each feature (e.g., word) usually not only appear to represent a single meaning, but also have correlations with others to reveal a complete and complex semantic. In comparison, the proposed approach adopts a Bi-directional LSTM network to effectively model the feature correlations for more accurate node relations modeling. The performance gain have demonstrated the benefits of the proposed model to learn accurate feature semantics.
    
        \item As can be seen from Table 2, Table 3 and Table4, models (e.g., GAT, FA-GCN$_{self}$ and FA-GCN) with incorporating either node-level attention mechanism or feature-level attention mechanism perform generally better than the basic GCN model. For example, on the Citeseer dataset, the average performance of GAT improved 0.9\% over GCN, and FA-GCN$_{self}$ and FA-GCN improved 2.2\% and 2.6\%, respectively. In general, edges in a graph could reveal complex relationships between nodes, \textit{i.e.}, in the citation network a paper may cite many others of various subject matters, and in the social network a user may connect many friends of different degrees of affinities. GCN enforces each node to indiscriminately aggregate information from all neighbors, which is inflexible and insufficient to model neighborhood relations between nodes. In comparison, GAT and the proposed attention model allow each node attends the important neighborhood nodes or their features for differentiable neighborhood features aggregation, which is helpful to accurately model node relations and meanwhile learn the alignment between the aggregated features and the node classification task. 

    \item In terms of methods with considering the attention models, from Table 2 and Table 4 we can observe that FA-GCN$_{self}$ outperforms GAT in most cases (e.g., the average performance improved 1.3\% w.r.t Citeseer dataset). The reason lies in that FA-GCN$_{self}$ adopts a more fine-grained attention mechanism at feature level, against GAT at node level. As we know, features in a node content could function differently, where two node sharing many identical features cannot guarantee they are highly similar. For example, in the citation network, the abstract of a publication has rich word features in which many are not distinguishing features to reveal the accurate topics involved. The node-level attention assumes that all features in each node content contribute equally, while the feature-level attention in FA-GCN$_{self}$ is able to assign higher weights to the most influential features for node representation learning and classification. In addition, as can be seen from all three result Tables, the proposed attention model FA-GCN can achieve even better performance than FA-GCN$_{self}$. The reason is because it has considered the contextual information while calculating each feature attention of all neighboring nodes, which enables each convolution node have more capacity in selecting the useful features aligned with the corresponding node classification. The experimental results have demonstrated the effectiveness of the proposed attention models.
\end{itemize}

\begin{figure}[t]
\begin{minipage}[b]{0.5\linewidth}
\centering 
\includegraphics[width=1\textwidth]{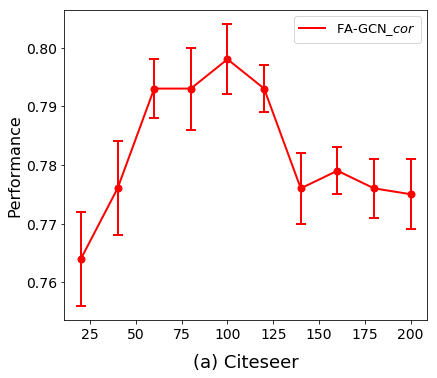}
\end{minipage}%
\begin{minipage}[b]{0.5\linewidth}
\centering 
\includegraphics[width=1\textwidth]{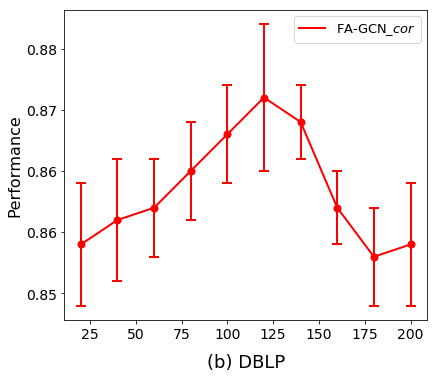}
\end{minipage}%
\caption{Impact of the input feature vector dimension $d_i$.}
\label{fig:5}
\vspace{-3mm}
\end{figure}

\begin{figure}[t]
\begin{minipage}[b]{0.5\linewidth}
\centering 
\includegraphics[width=1\textwidth]{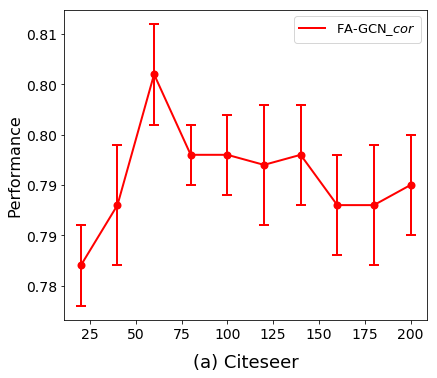}
\end{minipage}%
\begin{minipage}[b]{0.5\linewidth}
\centering 
\includegraphics[width=1\textwidth]{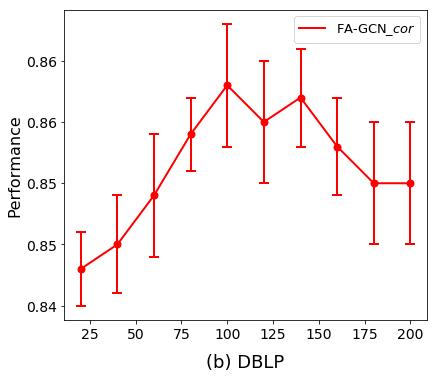}
\end{minipage}%
\caption{Impact of the output feature vector dimension $d_o$.}
\label{fig:5}
\end{figure}

\noindent \textbf{Noise Intervention Performance.} Figure 4 shows the performance of various GCN-based methods on all three graphs w.r.t different ratios of injected noisy features. As can be seen, with the increase of noise level, performances of all GCN-based methods such as GCN, GAT and FA-GCN$_{cor}$ tend to decline to some degrees. The phenomenon is mainly caused by the fact that in the spectral-based graph convolution learning process, nodes are forced to aggregate content features from their respective neighbors so as to maintain the complex link relationships in a graph. Once node contents are floated with noisy features, the content similarities between nodes would become less likely to reflect accurate neighborhood relations. Nevertheless, From Figure 4 (a) we can observe the proposed attention models are less sensitive to noises. The reason is that the introduced feature-level attention models are helpful to select the important features from the noisy, which has guaranteed, to some extent, the accurate neighborhood relations modeling. In addition, as can be seen Figure 4 (b) and (c), FA-GCN$_{self}$ and FA-GCN both outperform other baselines in most cases. Figure 5 shows the impact of the second type of noise intervention, which as can be seen has a significant impact on the performance of all comparative methods, \textit{i.e.}, with the noise ratio increased from 0.1 to 0.5, the performance of GCN decreased 3.4\% and 2.3\% on Citeseer and Cora, respectively. Nevertheless, the proposed FA-GCN model still performs better than other methods in most cases, which again demonstrated the effectiveness of the proposed models for sparse and noisy content network learning.
\vspace{0.1cm}

\begin{figure}[t]
\begin{minipage}[b]{0.5\linewidth}
\centering 
\includegraphics[width=1\textwidth]{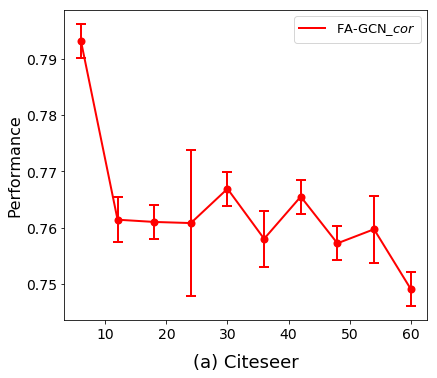}
\end{minipage}%
\begin{minipage}[b]{0.5\linewidth}
\centering 
\includegraphics[width=1\textwidth]{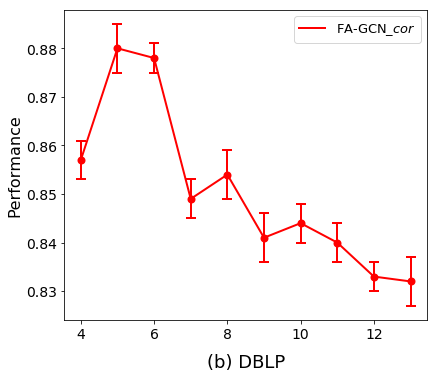}
\end{minipage}%
\caption{Impact of the hidden node vector dimension $d_h$.}
\label{fig:5}
\end{figure}

\noindent \textbf{Parameter Influence.} They are three hyper-parameters, $d_i$, $d_o$ and $d_h$, that are important in the feature and node representation learning process. Extensive experiments are designed to test their sensitivities on Citeseer and DBLP datasets. $d_i$ controls the dimension of the feature vector fed into the LSTM network. Figure 6 shows on both datasets the performance changes in a limited range which first increases and then drops with larger values pf $d_i$. $d_o$ indicates the dimension of the feature representation output by the LSTM network and its influences are shown in Figure 7. As can be seen, the performances first slightly increase and then decrease within a very small range after the dimensions are set as 60 and 100 for Citeseer and DBLP, respectively. Figure 8 shows the impact of parameter $d_h$ which represents the dimension of node representing out by the first convolutional layer in FA-GCN. We test impact of $d_h$ on Citeseer between 6 and 60, and on DBLP between 4 and 13. The results show it has a significant impact on the performance, where the performance dramatically declines after 6 and 5 for Citeseer and DBLP, respectively.

\section{Conclusions}
In this paper, we studied noise resilient learning for networks with sparse noisy node content. We argued that sparse, noisy, and erroneous graph content are ubiquitous. They present critical challenges to many graph learning methods that rely on network content to constrain and measure node relationships. To tackle feature sparsity, we first proposed to represent content features as dense vectors by an LSTM network, which leverages feature semantic correlation and dependency to learn dense vector for each feature. After that, we introduced a feature attention mechanism that allows each node to vary feature weight values with respect to different neighbors, allowing our method to minimize the noise impact and emphasize on consistent features between connected nodes. As a result, each node can gather the most important content features from itself and its neighbors to learn its node representation. The effectiveness of the proposed models have been validated on three sparse content benchmark networks. Experiments on noise-free and noisy networks, including different noise intervention by either injecting noise into the node content or replacing correct content features with error ones, confirm that the proposed method outperforms state-of-the-art methods such as GCN and GAT. Our method is less sensitive to erroneous graph contents, and is noise resilient for learning node representation.




\bibliographystyle{IEEEtran}
\bibliography{references}

\end{document}